\documentclass[a4paper]{article}

\usepackage{INTERSPEECH_v2}
\usepackage{amsmath,multirow,subcaption,graphicx,enumitem,caption}
\usepackage[ma]{}

\title{Using Lattice-based Features  for Out-of-Vocabulary Keywords Detection Task}

\title{Fast and Accurate OOV Decoder on High-Level Features}

\name{Yuri Khokhlov$^1$, Natalia Tomashenko$^{1,2,3}$, Ivan Medennikov$^{1,3}$, Alexei Romanenko$^{1,3}$}
\address{
  $^1$STC-Innovations Ltd,  Saint-Petersburg, Russia\\
  $^2$LIUM, University of Le Mans, France\\
  $^3$ITMO University, Saint-Petersburg, Russia}
\email{khokhlov@speechpro.com, natalia.tomashenko@univ-lemans.fr, medennikov@speechpro.com, romanenko@speechpro.com}

\begin{document}

\maketitle
\begin{abstract}

This work proposes a novel approach to out-of-vocabulary (OOV) keyword search
(KWS) task. 
The proposed approach is based on using high-level features from an automatic speech recognition (ASR) system, so called \textit{phoneme posterior based (PPB) } features, for decoding. These features are obtained by calculating time-dependent phoneme posterior probabilities from word lattices, followed by their smoothing. For the PPB features we developed a special novel very fast, simple and efficient OOV decoder. 
Experimental results are presented on the Georgian language from the IARPA Babel Program, which was the test language in the OpenKWS 2016 evaluation campaign. The results show that in terms of maximum term weighted value (MTWV) metric and computational speed, for single ASR systems, the proposed approach significantly outperforms the state-of-the-art approach based on using in-vocabulary proxies for OOV keywords in the indexed database. The comparison of the two OOV KWS approaches on the fusion results of the nine different ASR systems demonstrates that the proposed OOV decoder outperforms the proxy-based approach in terms of MTWV metric given the comparable processing speed. Other important advantages of the OOV decoder include extremely low memory consumption and simplicity of its implementation and parameter optimization.

\end{abstract}

\section{Introduction}



The keyword search (KWS) problem, which consists in finding a spoken or written word or a short word sequence in a collection of audio speech data, has remained  an active area of research during the last decade. Finding out-of-vocabulary (OOV) keywords -- those words, that are not known to the system  in advance at the training stage, is one of the fundamental  problem of KWS research.
Due to the growth of interest in development of low-resource speech recognition systems, the problem of OOV keyword search has become especially actual.

A variety of methods have been proposed in the literature to solve this problem.
Most of the state-of-the art KWS systems are based on the search in the indexed database. 
The speech indexing can be obtained from an automatic speech recognition (ASR) system output in the form of recognition
lattices or confusion networks (CNs) \cite{mangu2000finding}. There are two broad classes of methods for handling OOVs. 

The first class of methods is based
on representing OOVs with subword units \cite{szoke2008sub, hartmann2014comparing,bulyko2012subword,karakos2014subword,he2016using,yu2004hybrid,seide2004vocabulary,lee2016generating}, which can be used either in decoding stage  or obtained from the the word lattices. Various types of subword units have been explored in the literature, such as phones, graphones, syllables \cite{karakos2014subword}, morphes \cite{he2016using, hartmann2014comparing}, phone sequences of different length \cite{karakos2014subword} and charter n-grams \cite{hartmann2014comparing}.  Different types of  subword units have been shown to provide complementary results, so their combination (including word-level units) \cite{hartmann2014comparing, karakos2014subword}  and hybrid approaches \cite{yu2004hybrid} usually lead to further performance improvement. 

The second class of methods is based on using in-vocabulary (IV) \textit{proxies} -- those words that acoustically  are close to OOVs \cite{logan2002confusion, chen2013using, mangu2014efficient}. For this purpose confusion models are trained to expand the query \cite{logan2002confusion, saraclar2013empirical,mangu2014efficient, miller2007rapid, karanasou2012discriminatively, logan2005approaches} and perform fuzzy  search \cite{bulyko2012subword,karakos2014subword, miller2007rapid, mamou2008phonetic}.

This work proposes a novel approach to the OOV KWS task. 
The proposed approach is based on using 
\textit{phoneme posterior based (PPB) } features and a new decoding strategy for these features.
It was successfully used for the OpenKWS 2016 NIST evaluation campaign, as a part of the STC system~\cite{khokhlov2017thestc}.

The rest of the paper is organized as follows.  
In Section~\ref{sec:lbfeats}, PPB features are introduced.
The OOV decoder is presented in Section~\ref{sec:decoder}. 
Section~\ref{sec:results} describes the experimental results of the OOV KWS 
for the proposed approach and its comparison and combination with the proxy-based search.
Finally, the conclusions are presented in Section~\ref{sec:conclus}.

\section{Phoneme posterior based (PPB) features }
\label{sec:lbfeats}

In this section we present novel features which  are used in the proposed KWS system for OOV words. 
The extraction of the proposed PPB features  for audio files consists in the three major steps: 
\begin{enumerate}
\item Speech recognition;
\item Calculation of phoneme posterior probabilities from word lattices with phoneme alignments;
\item Smoothing of the obtained probabilities.
\end{enumerate}

\subsection{Calculation of phoneme posterior probabilities}

These PPB features are obtained from time dependent phoneme posterior scores \cite{uebel2001improvements, gollan2008confidence, evermann2000large} by computing arc posteriors from the output lattices of the decoder.
We use  the phone-level information from the lattices.
For each time frame we calculate $p_t^n$~--- the
confidence score of phoneme $ph_n$ at time $t$
 in the decoding lattice 
by calculating arc posterior probabilities. 
The forward-backward algorithm is used to calculate these arc posterior probabilities from the lattice as follows:

\begin{equation}
P(l|O)=\frac{\sum_{q\in Q_{l}}{p_{acc}(O|q)^{\frac{1}{\lambda}}P_{lm}(w)}}{P(O)},
\label{eq_poster}
\end{equation}
where $\lambda$ is the 
scale factor (the optimal value of $\lambda$ is found empirically by minimizing WER of the consensus hypothesis \cite{mangu2000finding});
$q$ is a path through the lattice corresponding to the word sequence $w$;
$Q_{l}$ is the set of paths passing through arc~$l$;
$p_{acc}(O|q)$ is the acoustic likelihood;
$P_{lm}(w)$ is the language model probability;
and $P(O)$ is the overall likelihood of all paths through the lattice.


Let $\left\{ph_{1},\ldots,ph_{N}\right\}$ be a set of phonemes including the silence model. 
For the given frame $\textbf{o}_t$ at time $t$ we calculate its probability $P(\textbf{o}_{t})\in ph_{n}$ of belonging to phoneme  $ph_n$, using  lattices obtained  from the first decoding pass:

\begin{equation}
p_t^n=P(\textbf{o}_{t}\in ph_{n})=\sum_{l\in S_{n}(\textbf{o}_{t})}{P(l|O)},
\label{eq_prob_for_weight}
\end{equation}
where $S_{n}(\textbf{o}_{t})$ is the set of all arcs corresponding to the phoneme $ph_n$ in the lattice at time $t$;
$P(l|O)$ is the posterior probability of arc $l$ in the lattice.

The obtained probability $P(\textbf{o}_{t}\in ph_{n})$ of frame $\textbf{o}_t$ belonging to phoneme $ph_n$ is the coordinate $p_t^n$ on the new feature vector $\textbf{p}_t$. 
Thus for a given acoustic feature vector $\textbf{o}_{t}$ at time $t$ we obtain a new feature vector $\textbf{p}_t$:
\begin{equation}\label{eq:p_t}
\textbf{p}_t=\left(p_t^{1},\ldots,p_t^N\right),
\end{equation}
where $N$ is the number of phones in the phoneme set used in the ASR system.

Hence for each frame $\textbf{o}_{t}$ we have a $N$-dimensional vector~$\textbf{p}_{t}$, each coordinate of which represents the probability of this frame to belong to a certain
phoneme.

\subsection{Smoothing}

The smoothing process consists of two steps:
\begin{enumerate}
\item Calculation of phoneme \textit{confusion model} $\textbf{M}$.
\item Transformation of vectors $\textbf{p}_t$ into \textit{smoothed vectors}  $\textbf{s}_t$ using the confusion model $\textbf{M}$.
\end{enumerate}

First, confusion model $\textbf{M}$ is calculated in an unsupervised manner on the development set from the decoding lattices as follows. It can be represented in the form of
$N\times N$ matrix:
$\textbf{M}=\left\{\ {\bm{\mu}}_1,\ldots,{\bm{\mu}}_N\right\}$,
 where ${\bm{\mu}}_n$ is the mean calculated over  all vectors  $\textbf{p}_t$,
 \unboldmath
 which "\textit{correspond}" to phoneme $ph_{n}$. The correspondence of vector   $\textbf{p}_t$ to phoneme $ph_{n}$
 means that:
 \begin{equation}
 \max_{k=1,\ldots,N}{p_t^k}=p_t^n.
 \end{equation}
In other words, 
 \begin{equation}
{\bm{\mu}}_n=\frac{1}{|T_n|}\sum_{t\in T_n}^{}{\textbf{p}_t},
 \end{equation}

\begin{equation}
T_n=\left\{t\in T: \operatorname*{arg\,max}_{k=1,\ldots,N}{p_t^k}=n  \right\}
 \end{equation}
 and $T$ is the set of all the frames  in the development set; $|T_n|$ is the number of elements in $T_n$.

In the second step, we perform smoothing  of 
$\textbf{p}_t$ vectors using the obtained confusion model as follows:

\begin{equation}\label{eq:s_t}
\textbf{s}_t=(1-\alpha)\textbf{p}_t+\alpha {\bm{\mu}}_n,
\end{equation}
where vector $\textbf{p}_t$ \textit{corresponds}
to phoneme $ph_n$.
The optimal value for $\alpha$ depends on lattice size and richness: the richer are the lattices the smaller can be the contribution of the confusion model. For the OOV decoder (described in Section~\ref{sec:decoder}) we use $\log(\textbf{s}_t)$ features.
When some phonemes are not present in the lattice for a certain frame, and when we also obtain "zero"  probabilities even after  the smoothing,
we set coordinates for them 
equal to some very small value  ($\epsilon\approx 10^{-42}$)  in  vector $\textbf{s}_t$.

\section{OOV decoder}\label{sec:decoder}
In this section we introduce the decoder developed for OOV search. We will refer to this decoder as \textit{OOV decoder}.
\subsection{Graph topology}
We use only a phone finite state automation (FSA) for decoding. 
This FSA is built for each OOV word independently and contains all possible pronunciation variants (according to phonetic transcriptions) of this word.
The other important  properties of topology of this FSA are as follows:
(1) it does not contain any loops; (2) no filler model is used.

Any loops in this FSA are not required because
the OOV decoder, which
works on PPB features,
attempts for each frame to generate  the hypothesis of the beginning of a keyword in this frame, if its probability exceeds a chosen threshold $\Theta_{start}$.

Also we do not need any filler or background models \cite{foote1997unconstrained, szoke2005comparison}, which are used, for example in acoustic KWS to absorb non-keyword speech events, because the OOV decoder works directly with probabilities and the decision to  accept or to reject a hypothesis is made on the basis of the final probability.

\subsection{Probability estimation and beam pruning}
Let $H$ be a current hypothesis of keyword $K$, which corresponds to 
phoneme sequence
$K=<\phi_1,\ldots,\phi_M>$.
Scores for this hypothesis are calculated in two steps:

\begin{enumerate}
\item Estimate probabilities of phones of the current hypothesis:
\begin{equation}
P(\phi_i)=\frac{1}{L_i}\sum_{t=t_{start(\phi_i)}}^{t_{end(\phi_i)}}{s_t^{\phi_i}},
\end{equation}
where $L_i=t_{end(\phi_i)}-t_{start(\phi_i)}+1$ is the length of the current phone  $\phi_i$ in the hypothesis $H$, 
$t_{start(\phi_i)}$ and $t_{end(\phi_i)}$ -- are the first and the final frames of phone  $\phi_i$ in $H$ correspondingly; 
$s_t^{\phi_i}$ -- is the $\phi_i$-th coordinate of vector $\textbf{s}_t$ (from Formula~(\ref{eq:s_t})). 

\item Estimate probability of the whole hypothesis $H$ as the average probability over all phones of the current keyword: 
\begin{equation}
P(H)=\frac{1}{M}\sum_{i=1}^{M}{P(\phi_i)}.
\end{equation}
\end{enumerate}

Note that this normalization of scores on the phoneme lengths allows us to compare  hypothesis of different lengths regardless of their duration.
It is a necessary because in the OOV decoder all hypothesis can have different lengths.

We reject a current hypothesis as soon as its probability becomes less the a given threshold $\Theta_{beam}$.
The maximum number of possible hypotheses in this decoder equals to the number of the FSA states.

\subsection{OOV keyword search}
For the OOV decoder described above the KWS becomes very simple:
if $P(H^{*})>\Theta_{hit}$ for a final hypothesis $H^{*}$, then we accept this keyword and add it to the keyword list with the corresponding score $\log\left(P(H^{*})\right)$. Here $\Theta_{hit}$  is the constant threshold fixed for all keywords.
In the end we apply the sum-to-one (STO) \cite{mangu2013exploiting,mamou2013system} score normalization to all keyword queries.

\section{Experimental results}\label{sec:results}

\subsection{Training and test data}
Experiments were performed on the Georgian language 
from the IARPA Babel Program, which was the "surprise" language in the OpenKWS 2016 evaluation campaign.
For acoustic model  training 40 hours of transcribed and 40 hours of untranscribed data were used.
Results presented in this paper are reported for the official development  set (10 hours).
Additional data from 18 other Babel languages with the total amount of 860 hours were used to train a multilingual feature extractor.

\subsection{ASR system}

\subsubsection{Acoustic models}

We used the Kaldi speech recognition toolkit~\cite{povey2011kaldi}
for AM training (with some additional modifications) and for decoding. Nine different  neural network (NN) acoustic models (AMs) were used in these experiments. They differ in type, topology, input features, training data and learning algorithms. 
The detailed description of the AMs is given in \cite{khokhlov2017thestc}.
Here we only listed the main points.

First, two multi-lingual (ML) NN models  were trained:  (1)~deep NN (DNN) for speaker-dependent (SD) bottleneck (BN) features with i-vectors; (2) deep maxout network (DMN) for SD-BN features with i-vectors.
Then the nine final AMs were trained on the training dataset for the Georgian language:

\begin{enumerate}[leftmargin=*]

 \item \textbf{DNN}$_1$ is a sequence-trained DNN  with a state-level Minimum Bayes Risk (sMBR) criterion  on 11$\times$(perceptual linear predictive (PLP) $+$ pitch features, adapted using feature space maximum likelihood linear regression (fMLLR) adaptation);

\item \textbf{DNN}$_2$ is a DNN  trained with sMBR on  31$\times$(fMLLR-adapted SD-BN features from ML DNN);

\item \textbf{DMN}$_3$ is a DMN trained with sMBR on 31$\times$(SD-BN features from ML DMN);

\item \textbf{DMN}$_4$ is similar to DMN$_3$, but initialized with a shared part of ML DMN;

\item \textbf{TDNN}$_5$ is a time delay neural network (TDNN) trained as described in \cite{peddinti2015time};

\item \textbf{BLSTM}$_6$ is a bidirectional long short-term memory (BLSTM)  network trained with  cross-entropy (CE) criterion on 5$\times$(fbank$+$pitch) features with i-vectors;

\item \textbf{DNN}$_7$ is a CE-trained DNN on 11$\times$ (PLP$+$pitch) features  with i-vectors; initialization with shared part of ML  DNN;

\item \textbf{DMN}$_8$ is a CE-trained DMN on 11$\times$ (fbank $+$ pitch) features with i-vectors; initialization with the shared part of ML DMN;

\item \textbf{DMN}$_9$ is similar to  DMN$_8$, but with  semi-supervised learning on the additional untranscribed part of the dataset.

\end{enumerate}

All the above AMs except the first one were trained with the use of speed perturbed  data~\cite{ko2015audio}.
The performance results for these models (with the  language model (LM) described in Section~\ref{sec:lm}) on the development set in terms of word error rate (WER)  are reported in Table~\ref{tab:all_models}.

\begin{table}[h]
	\caption{\label{tab:oov-dec-1} {\it Comparison of two approaches for OOV KWS
    in terms of MTWV metric and speed for different AMs}}
    \label{tab:all_models}
    \centerline{\renewcommand{\tabcolsep}{1.15mm}
        \centering
            \begin{tabular}{l|c|cc|cc}
            \toprule
             &  & \multicolumn{2}{c|}{\textbf{OOV decoder}} &     \multicolumn{2}{|c}{\textbf{Proxies}}   \\ 
            \textbf{AM} & \textbf{WER},\textbf{\%} & \textbf{MTWV}  & \textbf{RTF} & \textbf{MTWV} & \textbf{RTF}   \\ 
                \midrule 
DNN$_1$	&	44.2	&	0.561	&	6.2e-05	&	0.440	&	0.0016\\
DNN$_2$	&	41.5	&	0.548	&	5.7e-05	&	0.449	&	0.0015\\
DMN$_3$	&	39.4	&	0.591	&	5.7e-05	&	0.512	&	0.0014\\
DMN$_4$	&	44.3	&	0.579	&	5.7e-05	&	0.492	&	0.0015\\
TDNN$_5$	&	42.3	&	0.579	&	5.7e-05	&	0.490	&	0.0015\\
BLSTM$_6$	&	41.1	&	0.559	&	5.8e-05	&	0.537	&	0.0015\\
DNN$_7$	&	43.0	&	0.586	&	5.8e-05	&	0.517	&	0.0015\\
DMN$_8$	&	42.4	&	0.615	&	5.8e-05	&	0.491	&	0.0017\\
DMN$_9$	&	41.8	&	0.630	&	5.8e-05	&	0.528	&	0.0015\\
\hline
	        \end{tabular}
            }
\end{table}

\subsubsection{Language modeling}
\label{sec:lm}
The LM used in these experiments was obtained as a linear interpolation of the three LMs: (1) baseline LM; (2) LM-char; (3) LM-web. These  trigram LMs were trained with the SRILM toolkit~\cite{stolcke2002srilm}.
The first (baseline) LM was trained on transcriptions of the 40 hours of the FullLP dataset.
The second (LM-char) LM was trained 
on artificially generated text data  by a character based recurrent neural network (Char-RNN) LM using~\cite{karpathy2015unreasonable}.
The third LM (LM-web) was trained on extra data (web texts, about 380 Mb) provided by the organizers (BBN part). 
The size of the lexicon for LMs trained on artificial and web texts was limited to 150K. More details about the LMs are provided in \cite{khokhlov2017thestc}.

\subsection{Baseline KWS system: using IV proxies for OOVs}

For comparison purpose with the proposed approach we took as a baseline method one of the most efficient  algorithms, developed for OOV KWS  in the indexed database -- OOV search with proxies \cite{chen2013using, mangu2014efficient}.
%
In our KWS implementation we used a word-level CN \cite{mangu2014efficient} based index for  OOV search. 
The idea is based on the \textit{proxy-based approach} proposed in \cite{mangu2014efficient}, where a special weighted finite-state transducer (WFST) is constructed from a CN, and used to search for IV and OOV words. 
We applied several modifications (described in \cite{khokhlov2017thestc}) to the original algorithm~\cite{mangu2014efficient} in order to speed up the search process and to improve the performance.

\subsection{Results}

Due to the use of Char-RNN model for text generation in LM training, we have a low number of OOVs remained in the development set (only 93 OOVs left from the official keyword list).    
For this reason  we artificially created an additional OOV list,  by using a procedure
described in \cite{cui2014automatic}, 
in order to perform more representative experiments. The resulting total number of OOVs is 742. The number  of OOV targets in the development set is 796. 
The performance of the systems is evaluated using the Maximum Term-Weighted Value (MTWV) metric \cite{fiscus2006spoken}.

\begin{figure}[t]
\vspace{-0.5cm}
  \centering
  \includegraphics[width=0.98\linewidth]{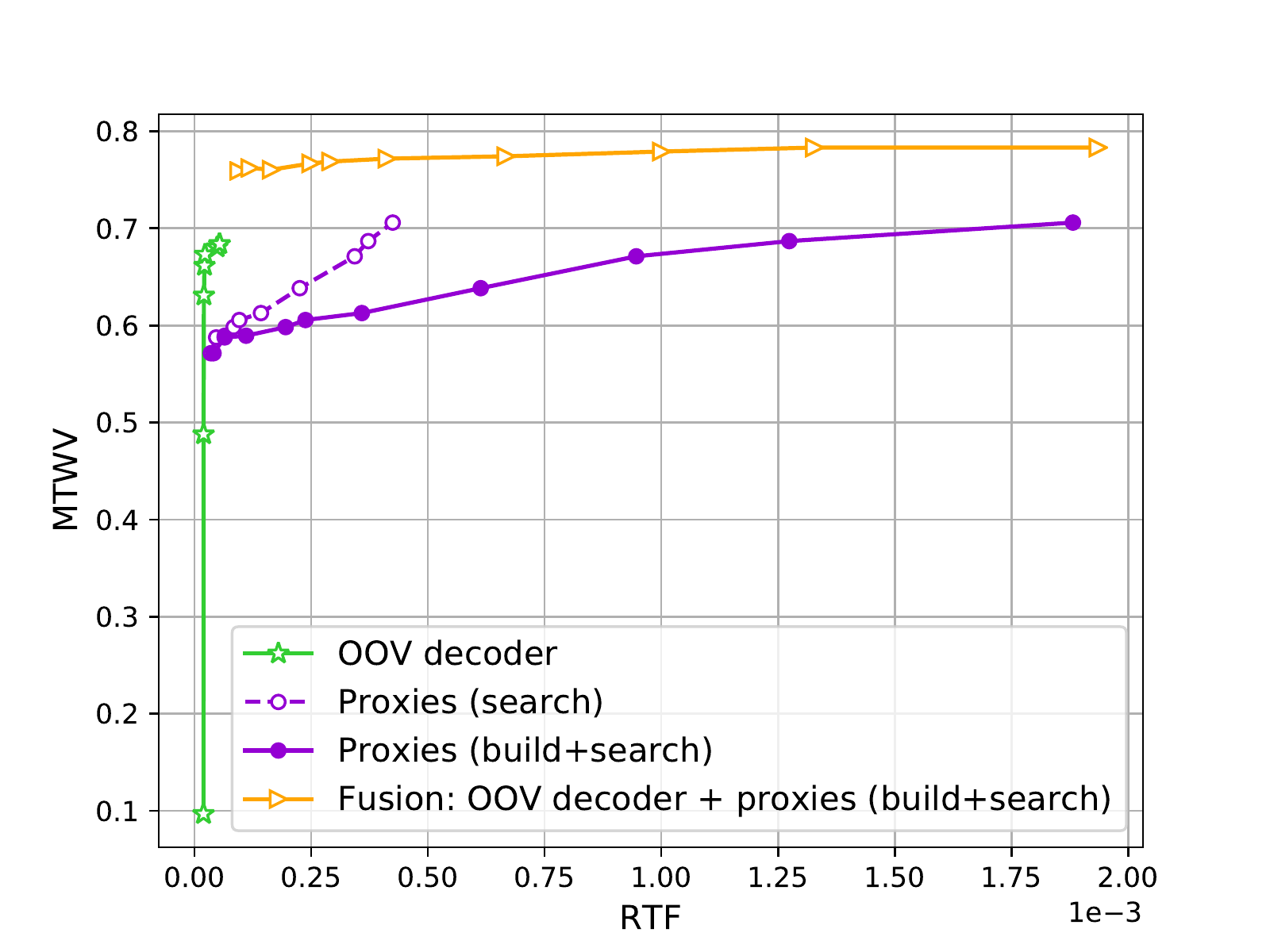}
  \caption{Comparison of two approaches for OOV KWS:  (1)~OOV decoder and (2) proxies
and their fusion  
for results from the fusion of the 9 ASR systems} 
  \label{fig:oov1}
  \vspace{-0.3cm}
\end{figure}

Experimental results
show that the proposed OOV decoder significantly outperforms (in terms of MTWV metric and speed) the approach based on using proxies for all AMs (Table~\ref{tab:all_models}) and provides 4--28\% of relative MTWV improvement for different AMs. The real time factor (RTF) was calculated per word. 
The comparison of the two OOV KWS approaches on the lattice-level fusion results of the nine different ASR systems demonstrates (Figure~\ref{fig:oov1}) that the  OOV decoder significantly outperforms the proxy-based search in terms of MTWV metric given the comparable processing speed.
For proxies, not only the search speed is important, but also the speed of their WFST generation (denoted as 'build' in Figure~\ref{fig:oov1}), which becomes especially crucial when it is required to search for a large number of keywords in a relatively small database.
Figure~\ref{fig:oov1} also demonstrates a very effective strategy for OOV KWS: fusion of the OOV decoder and the proxy-based search (in the point  of low RTF) can provide a very fast search method that significantly  outperforms the both approaches in MTWV.

In addition to the lattice-level fusion, we performed fusion on the list-level, 
described in \cite{trmal2014keyword},
using Kaldi \cite{povey2011kaldi} for all AMs for each approach independently and  for the both approaches together (Table~\ref{tab:fus}).
The list-level combination of all the systems for both approaches provides an additional improvement in overall accuracy (MTWV=0.795), which corresponds to 7.4\% of relative MTWV improvement over the best fusion result.

\begin{figure}[h]
\vspace{-0.5cm}
\begin{center}
  \begin{subfigure}{0.46\textwidth}
  \includegraphics[width=\linewidth]{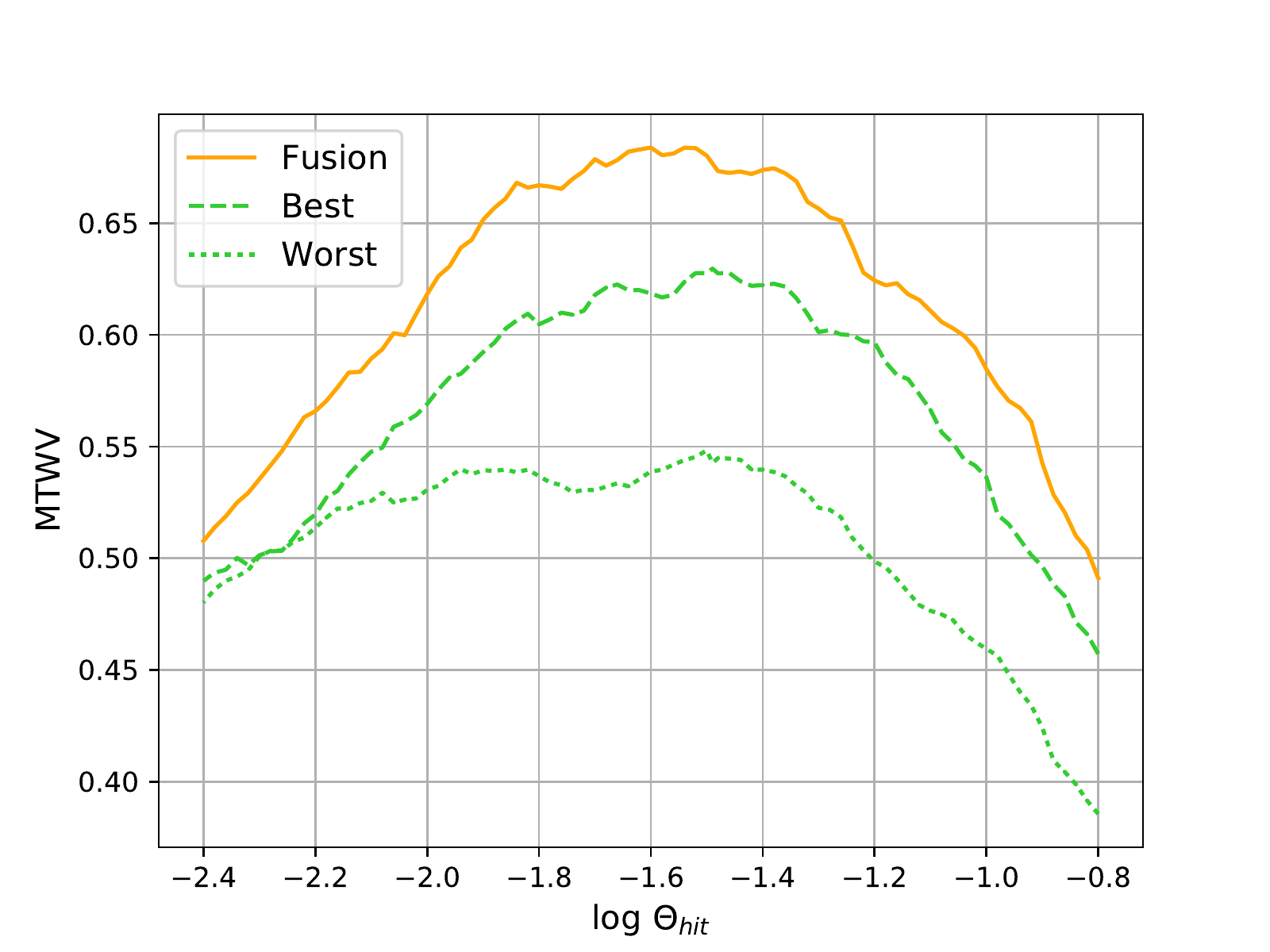}
  \caption{}
  
  \label{fig:minth}
  \vspace{-0.1cm}
\end{subfigure}
\begin{subfigure}{0.46\textwidth}
  \centering
  \includegraphics[width=\linewidth]{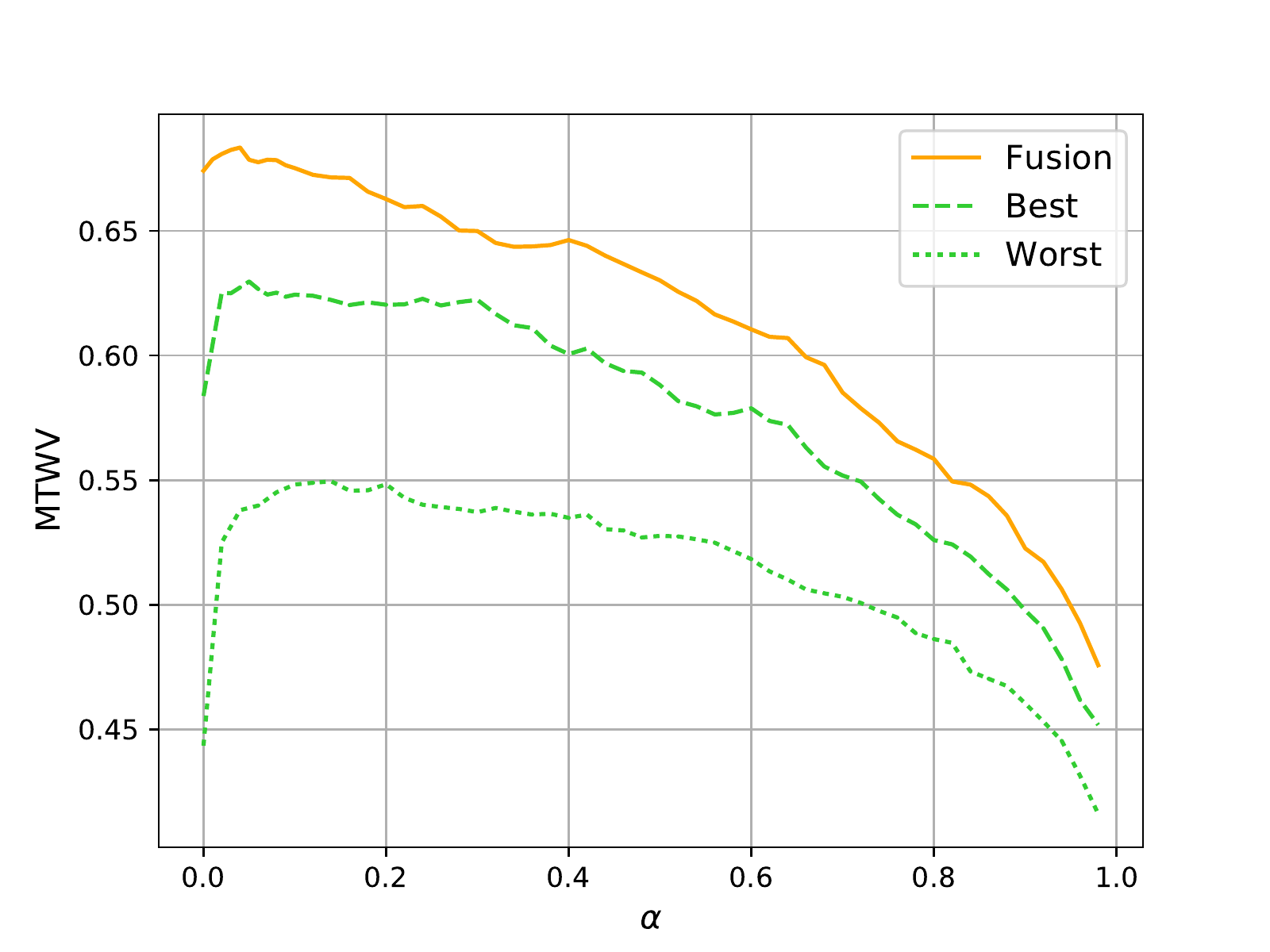}
  \caption{}
  \label{fig:smoothw}
\end{subfigure}
 \caption{Dependence of the KWS accuracy for OOV decoder on parameters: (a) $\Theta_{hit}$ -- minimum score threshold, and (b)~$\alpha$~ -- smoothing weight} 
\vspace{-0.82cm}
\end{center}
\end{figure}

Exploration of the two main parameters of the OOV decoder (smoothing weight $\alpha$ and minimum score threshold $\Theta_{hit}$) is presented in Figures~\ref{fig:minth} and \ref{fig:smoothw}.
The results are given for the best AM (DNN$_2$), for the worst AM (DMN$_9$) and for the lattice-based fusion of all AMs.

\begin{table}[h]
	\caption{\label{tab:fus} {\it Fusion results approaches for two OOV KWS approaches in terms of MTWV metric}}
    \label{tab:fus}
        \centering
            \begin{tabular}{lcc}
            \toprule
           \textbf{Fusion}   & \textbf{OOV decoder} &     \textbf{Proxies}     \\  \midrule 
           Lattices          & 0.684 & 0.706 \\   
           Lists             & 0.740 & 0.682 \\ \toprule
           \textbf{Lists     (for all systems) }      & \multicolumn{2}{c}{\textbf{0.795}} \\
%
%
\hline
	        \end{tabular}
\end{table}


\section{Conclusions}\label{sec:conclus}
We have presented a novel approach for OOV keywords detection. 
This approach utilizes the phoneme posterior based  features for decoding.
Experimental results have  demonstrated that in terms of MTWV metric and computational speed, for single ASR systems and for the list-level fusion of the results obtained from multiple ASR systems, the proposed algorithm significantly outperforms the proxy-based approach and provides in average 18.1\% of relative MTWV improvement.
The combination of the OOV decoder and proxy-based search provides an additional gain  of 12.6\%
 relative MTWV improvement over the best result obtained from the fusion results for the proxy-based method. 
 The OOV decoder works 23-43 times faster than the proxy-based search in the point of comparable MTWV values.
We have found that the combination of the two KWS methods provides an effective search strategy -- a high KWS accuracy can be reached with a low RTF.
The analysis of the parameters of the OOV decoder 
shows that the feature smoothing allows us to significantly improve the accuracy of OOV word search, what is especially important in the case of sparse lattices. Also the minimum score threshold parameter has a strong impact on MTWV and has to be optimized.
In comparison with the proxy-based search 
 the OOV decoder requires extremely low memory consumption and is very simple in implementation and parameter optimization.

\section{Acknowledgements}


The work was financially supported by the Ministry of Education and Science of the Russian Federation. Contract 14.579.21.0121, ID RFMEFI57915X0121.

This effort uses the IARPA Babel Program language collection release IARPA-babel\{101b-v0.4c, 102b-v0.5a, 103b-v0.4b, 201b-v0.2b, 203b-v3.1a, 205b-v1.0a, 206b-v0.1e, 207b-v1.0e, 301b-v2.0b, 302b-v1.0a, 303b-v1.0a, 304b-v1.0b, 305b-v1.0c, 306b-v2.0c, 307b-v1.0b, 401b-v2.0b, 402b-v1.0b, 403b-v1.0b, 404b-v1.0a\}, set of training transcriptions and BBN part of clean web data for Georgian language.

\bibliographystyle{IEEEtran}



\bibliography{mybib}

\begin{thebibliography}{10}
\providecommand{\url}[1]{#1}
\csname url@samestyle\endcsname
\providecommand{\newblock}{\relax}
\providecommand{\bibinfo}[2]{#2}
\providecommand{\BIBentrySTDinterwordspacing}{\spaceskip=0pt\relax}
\providecommand{\BIBentryALTinterwordstretchfactor}{4}
\providecommand{\BIBentryALTinterwordspacing}{\spaceskip=\fontdimen2\font plus
\BIBentryALTinterwordstretchfactor\fontdimen3\font minus
  \fontdimen4\font\relax}
\providecommand{\BIBforeignlanguage}[2]{{%
\expandafter\ifx\csname l@#1\endcsname\relax
\typeout{** WARNING: IEEEtran.bst: No hyphenation pattern has been}%
\typeout{** loaded for the language `#1'. Using the pattern for}%
\typeout{** the default language instead.}%
\else
\language=\csname l@#1\endcsname
\fi
#2}}
\providecommand{\BIBdecl}{\relax}
\BIBdecl

\bibitem{mangu2000finding}
L.~Mangu, E.~Brill, and A.~Stolcke, ``Finding consensus in speech recognition:
  word error minimization and other applications of confusion networks,''
  \emph{Computer Speech \& Language}, vol.~14, no.~4, pp. 373--400, 2000.

\bibitem{szoke2008sub}
I.~Szoke, L.~Burget, J.~Cernocky, and M.~Fapso, ``Sub-word modeling of out of
  vocabulary words in spoken term detection,'' in \emph{2008 IEEE Spoken
  Language Technology Workshop}, Dec 2008, pp. 273--276.

\bibitem{hartmann2014comparing}
W.~Hartmann, V.~B. Le, A.~Messaoudi, L.~Lamel, and J.-L. Gauvain, ``Comparing
  decoding strategies for subword-based keyword spotting in low-resourced
  languages.'' in \emph{Interspeech}, 2014, pp. 2764--2768.

\bibitem{bulyko2012subword}
I.~Bulyko, J.~Herrero, C.~Mihelich, and O.~Kimball, ``Subword speech
  recognition for detection of unseen words,'' in \emph{Thirteenth Annual
  Conference of the International Speech Communication Association,
  Interspeech}, 2012.

\bibitem{karakos2014subword}
D.~Karakos and R.~M. Schwartz, ``Subword and phonetic search for detecting
  out-of-vocabulary keywords.'' in \emph{Interspeech}, 2014, pp. 2469--2473.

\bibitem{he2016using}
Y.~He, P.~Baumann, H.~Fang, B.~Hutchinson, A.~Jaech, M.~Ostendorf,
  E.~Fosler-Lussier, and J.~Pierrehumbert, ``Using pronunciation-based
  morphological subword units to improve {OOV} handling in keyword search,''
  \emph{IEEE/ACM Transactions on Audio, Speech, and Language Processing},
  vol.~24, no.~1, pp. 79--92, Jan 2016.

\bibitem{yu2004hybrid}
P.~Yu and F.~T.~B. Seide, ``A hybrid word/phoneme-based approach for improved
  vocabulary-independent search in spontaneous speech.'' in
  \emph{Interspeech}.\hskip 1em plus 0.5em minus 0.4em\relax Citeseer, 2004.

\bibitem{seide2004vocabulary}
F.~Seide, P.~Yu, C.~Ma, and E.~Chang, ``Vocabulary-independent search in
  spontaneous speech,'' in \emph{Acoustics, Speech, and Signal Processing,
  2004. Proceedings.(ICASSP'04). IEEE International Conference on},
  vol.~1.\hskip 1em plus 0.5em minus 0.4em\relax IEEE, 2004, pp. I--253.

\bibitem{lee2016generating}
S.-w. Lee, K.~Tanaka, and Y.~Itoh, ``Generating complementary acoustic model
  spaces in {DNN}-based sequence-to-frame {DTW} scheme for out-of-vocabulary
  spoken term detection,'' \emph{Interspeech 2016}, pp. 755--759, 2016.

\bibitem{logan2002confusion}
B.~Logan and J.-M. Van~Thong, ``Confusion-based query expansion for oov words
  in spoken document retrieval.'' in \emph{Interspeech}, 2002.

\bibitem{chen2013using}
G.~Chen, O.~Yilmaz, J.~Trmal, D.~Povey, and S.~Khudanpur, ``Using proxies for
  oov keywords in the keyword search task,'' in \emph{Automatic Speech
  Recognition and Understanding (ASRU), 2013 IEEE Workshop on}.\hskip 1em plus
  0.5em minus 0.4em\relax IEEE, 2013, pp. 416--421.

\bibitem{mangu2014efficient}
L.~Mangu, B.~Kingsbury, H.~Soltau, H.-K. Kuo, and M.~Picheny, ``Efficient
  spoken term detection using confusion networks,'' in \emph{Acoustics, Speech
  and Signal Processing (ICASSP), 2014 IEEE International Conference on}.\hskip
  1em plus 0.5em minus 0.4em\relax IEEE, 2014, pp. 7844--7848.

\bibitem{saraclar2013empirical}
M.~Saraclar, A.~Sethy, B.~Ramabhadran, L.~Mangu, J.~Cui, X.~Cui, B.~Kingsbury,
  and J.~Mamou, ``An empirical study of confusion modeling in keyword search
  for low resource languages,'' in \emph{2013 IEEE Workshop on Automatic Speech
  Recognition and Understanding}, Dec 2013, pp. 464--469.

\bibitem{miller2007rapid}
D.~R. Miller, M.~Kleber, C.-L. Kao, O.~Kimball, T.~Colthurst, S.~A. Lowe, R.~M.
  Schwartz, and H.~Gish, ``Rapid and accurate spoken term detection.''

\bibitem{karanasou2012discriminatively}
P.~Karanasou, L.~Burget, D.~Vergyri, M.~Akbacak, and A.~Mandal,
  ``Discriminatively trained phoneme confusion model for keyword spotting.'' in
  \emph{Interspeech}, 2012, pp. 2434--2437.

\bibitem{logan2005approaches}
B.~Logan, J.-M. Van~Thong, and P.~J. Moreno, ``Approaches to reduce the effects
  of oov queries on indexed spoken audio,'' \emph{IEEE transactions on
  multimedia}, vol.~7, no.~5, pp. 899--906, 2005.

\bibitem{mamou2008phonetic}
J.~Mamou and B.~Ramabhadran, ``Phonetic query expansion for spoken document
  retrieval.'' in \emph{Interspeech}, 2008, pp. 2106--2109.

\bibitem{khokhlov2017thestc}
Y.~Khokhlov, I.~Medennikov, A.~Romanenko, V.~Mendelev, M.~Korenevsky,
  A.~Prudnikov, N.~Tomashenko, and A.~Zatvornitsky, ``The {STC} keyword search
  system for {OpenKWS} 2016 evaluation.'' in \emph{Interspeech}, 2017.

\bibitem{uebel2001improvements}
L.~Uebel and P.~C. Woodland, ``Improvements in linear transform based speaker
  adaptation,'' in \emph{Proc. ICASSP}, 2001, pp. 49--52.

\bibitem{gollan2008confidence}
C.~Gollan and M.~Bacchiani, ``Confidence scores for acoustic model
  adaptation,'' in \emph{Proc. ICASSP}, 2008, pp. 4289--4292.

\bibitem{evermann2000large}
G.~Evermann and P.~C. Woodland, ``Large vocabulary decoding and confidence
  estimation using word posterior probabilities,'' in \emph{Acoustics, Speech,
  and Signal Processing, 2000. ICASSP'00. Proceedings. 2000 IEEE International
  Conference on}, vol.~3.\hskip 1em plus 0.5em minus 0.4em\relax IEEE, 2000,
  pp. 1655--1658.

\bibitem{foote1997unconstrained}
J.~Foote, S.~J. Young, G.~J. Jones, and K.~S. Jones, ``Unconstrained keyword
  spotting using phone lattices with application to spoken document
  retrieval,'' \emph{Computer Speech \& Language}, vol.~11, no.~3, pp.
  207--224, 1997.

\bibitem{szoke2005comparison}
I.~Sz{\"o}ke, P.~Schwarz, P.~Matejka, L.~Burget, M.~Karafi{\'a}t, M.~Fapso, and
  J.~Cernock{\`y}, ``Comparison of keyword spotting approaches for informal
  continuous speech.'' in \emph{Interspeech}.\hskip 1em plus 0.5em minus
  0.4em\relax Citeseer, 2005, pp. 633--636.

\bibitem{mangu2013exploiting}
L.~Mangu, H.~Soltau, H.~K. Kuo, B.~Kingsbury, and G.~Saon, ``Exploiting
  diversity for spoken term detection,'' in \emph{2013 IEEE International
  Conference on Acoustics, Speech and Signal Processing}, May 2013, pp.
  8282--8286.

\bibitem{mamou2013system}
J.~Mamou, J.~Cui, X.~Cui, M.~J. Gales, B.~Kingsbury, K.~Knill, L.~Mangu,
  D.~Nolden, M.~Picheny, B.~Ramabhadran \emph{et~al.}, ``System combination and
  score normalization for spoken term detection,'' in \emph{Acoustics, Speech
  and Signal Processing (ICASSP), 2013 IEEE International Conference on}.\hskip
  1em plus 0.5em minus 0.4em\relax IEEE, 2013, pp. 8272--8276.

\bibitem{povey2011kaldi}
D.~Povey, A.~Ghoshal, G.~Boulianne, L.~Burget, O.~Glembek, N.~Goel,
  M.~Hannemann, P.~Motlicek, Y.~Qian, P.~Schwarz \emph{et~al.}, ``The kaldi
  speech recognition toolkit,'' in \emph{IEEE 2011 workshop on automatic speech
  recognition and understanding}, no. EPFL-CONF-192584.\hskip 1em plus 0.5em
  minus 0.4em\relax IEEE Signal Processing Society, 2011.

\bibitem{peddinti2015time}
V.~Peddinti, D.~Povey, and S.~Khudanpur, ``A time delay neural network
  architecture for efficient modeling of long temporal contexts.'' in
  \emph{Interspeech}, 2015, pp. 3214--3218.

\bibitem{ko2015audio}
T.~Ko, V.~Peddinti, D.~Povey, and S.~Khudanpur, ``Audio augmentation for speech
  recognition.'' in \emph{Interspeech}, 2015, pp. 3586--3589.

\bibitem{stolcke2002srilm}
A.~Stolcke, ``{SRILM-an} extensible language modeling toolkit.'' in \emph{7th
  International Conference on Spoken Language Processing, ICSLP}, vol. 2002,
  2002, p. 2002.

\bibitem{karpathy2015unreasonable}
A.~Karpathy, ``The unreasonable effectiveness of recurrent neural networks,''
  \emph{http://karpathy.github.io/2015/05/21/rnn-effectiveness}, 2015.

\bibitem{cui2014automatic}
J.~Cui, J.~Mamou, B.~Kingsbury, and B.~Ramabhadran, ``Automatic keyword
  selection for keyword search development and tuning,'' in \emph{Acoustics,
  Speech and Signal Processing (ICASSP), 2014 IEEE International Conference
  on}.\hskip 1em plus 0.5em minus 0.4em\relax IEEE, 2014, pp. 7839--7843.

\bibitem{fiscus2006spoken}
J.~Fiscus, J.~Ajot, and G.~Doddington, ``The spoken term detection ({STD}) 2006
  evaluation plan,'' \emph{NIST USA, Sep}, 2006.

\bibitem{trmal2014keyword}
J.~Trmal, G.~Chen, D.~Povey, S.~Khudanpur, P.~Ghahremani, X.~Zhang, V.~Manohar,
  C.~Liu, A.~Jansen, D.~Klakow \emph{et~al.}, ``A keyword search system using
  open source software,'' in \emph{Spoken Language Technology Workshop (SLT),
  2014 IEEE}.\hskip 1em plus 0.5em minus 0.4em\relax IEEE, 2014, pp. 530--535.

\end{thebibliography}

\end{document}